\documentclass[10pt, conference]{ieeeconf}      

\IEEEoverridecommandlockouts   

\usepackage{graphics} 
\usepackage{epsfig} 
\usepackage{times} 
\usepackage{amsmath} 
\usepackage{amssymb}  
 \usepackage{url}
 \usepackage{color}
 \usepackage{subfig}
 \usepackage{authblk}

\newenvironment{definition}[1][Definition]{\begin{trivlist}
\item[\hskip \labelsep {\bfseries #1}]}{\end{trivlist}}

\title{\LARGE \bf Autism Spectrum Disorder Classification using Graph Kernels on Multidimensional Time Series}
\author{Rushil Anirudh}
\author{Jayaraman J. Thiagarajan \thanks{This work was performed under the auspices of the U.S. Dept. of Energy by Lawrence Livermore National Laboratory under Contract DE-AC52-07NA27344.}}
\affil{Lawrence Livermore National Laboratory \\ Email: \{anirudh1, jjayaram\}@llnl.gov}

\author{Irene Kim}
\author{Wolfgang Polonik}
\affil{University of California, Davis \\ Email: \{imkkim, wpolonik\}@ucdavis.edu}

\begin{document}

\maketitle
\thispagestyle{empty}
\pagestyle{empty}

\begin{abstract}
We present an approach to model time series data from resting state fMRI for autism spectrum disorder (ASD) severity classification. We propose to adopt kernel machines and employ graph kernels that define a kernel dot product between two graphs. This enables us to take advantage of spatio-temporal information to capture the dynamics of the brain network, as opposed to aggregating them in the spatial or temporal dimension. In addition to the conventional similarity graphs, we explore the use of $\ell_1$ graph using sparse coding, and the persistent homology of time delay embeddings, in the proposed pipeline for ASD classification. In our experiments on two datasets from the ABIDE collection, we demonstrate a consistent and significant advantage in using graph kernels over traditional linear or non linear kernels for a variety of time series features. 
\end{abstract}

\section{Introduction}
\label{sec:intro}
Modeling the relationships between functional or structural regions in the brain is a significant step towards understanding, diagnosing and eventually treating a gamut of neurological conditions including epilepsy, stroke, and autism. Sensing mechanisms such as functional-MRI, Electrocorticography (ECoG), and Electroencephalography (EEG) can record information related to brain activity. In order to truly model the dynamics of the brain network, one has to consider the spatio-temporal properties of the signals -- i.e. understanding how different regions encode information and interact with each other, over time. 

We consider the problem of classifying Autism Spectrum Disorder (ASD), which is a group of developmental disorders. Recent collaborative efforts such as the Autism Brain Imaging Data Exchange (ABIDE) \cite{ABIDEpaper2014}, are seeking to bring a \emph{big data} approach towards understanding and diagnosing ASD. Clinical measures such as the the Autism Diagnostic Observation Schedule (ADOS), quantify the severity of the condition in a subject, and our goal is to predict the severity directly from resting state fMRI data. What makes ASD prediction particularly hard is that the data from resting state fMRI is significantly more challenging when compared to similar data from other task-based studies. Furthermore, researchers have reported that the ADOS score can be biased heavily by the subject's developmental and language levels \cite{Moradi2016}. As a result, predictive models learned on features from the time series data generalize poorly. In addition, considering multiple locations within the fMRI requires the design of spatial or temporal aggregation strategies. Despite these inherent challenges, predictive modeling of ASD has received increased research interest, due to the availability of larger datasets. For example, Sato et al. \cite{sato2013inter} showed that there is a relationship between inter-regional cortical thickness and autism symptoms, using regression studies. More recently, Moradi et al. \cite{Moradi2016} proposed to predict the severity using cortical thickness measurements from the ABIDE dataset. An interesting aspect of their work is in using domain adaptation to map data from different sites, into a more homogeneous space for inference. 

In this paper, we develop an alternative approach that utilizes kernel similarities on graphs, which can encode both spatial as well as the temporal characteristics. Graphs are a natural representation to study both the structural and functional properties of the brain network \cite{bullmore2009complex}. Several recent studies have explored graphs to analyze fMRI data \cite{gkirtzou2013fmri,jie2014topological}. Despite being crucial to constructing meaningful graphs, the problem of choosing the appropriate similarity metric has not been addressed so far. We build a flexible prediction pipeline based on graph kernels and explore the impact of different graph construction strategies on the prediction performance. In addition to commonly adopted correlation and $\ell_2$-norm based graphs, we employ sparse coding based $\ell_1$ graphs and persistence homology-based construction. While the former utilizes a generative model (union of subspaces) to describe the complex structure, the latter approach builds time-delay embeddings and extracts topological features for comparing the time series from different sites in the brain. Note that our approach is different from the topological analysis reported in \cite{WongISBI2016}, which constructs a simplicial complex for each patient based on correlations and extracts its persistence homology for regression based on persistence space kernels \cite{NIPS2015_5887}. Using empirical studies with the ABIDE dataset, we show conclusively that the proposed prediction pipeline performs consistently better than traditional methods. Furthermore, we evaluate the performance of different graph construction strategies using the proposed approach and present discussions on their merits and limitations.


\section{Background}
\label{sec:background}
\begin{figure*}[t]
\centering
\includegraphics[trim={0 142 0 103},clip,width=5.5in]{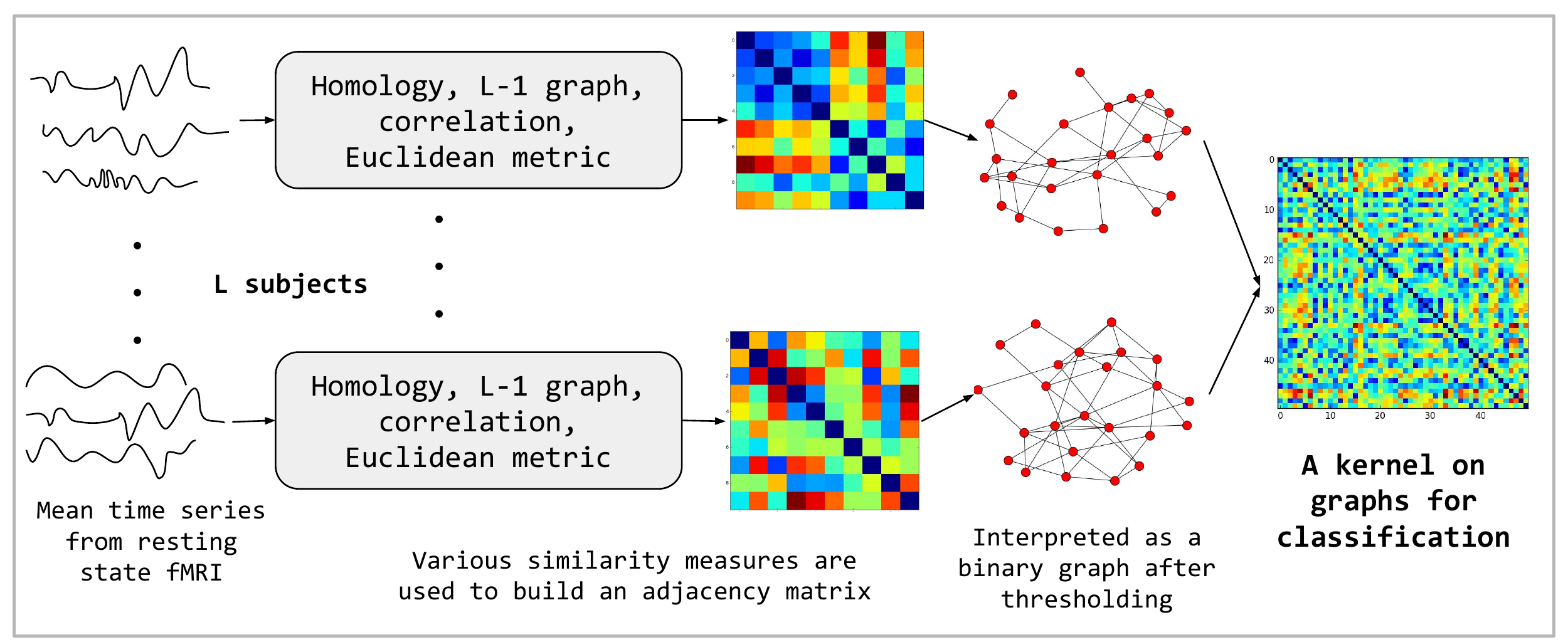}
\label{fig:overview}
\caption{\small{\bf An overview of the proposed system. We encode the similarities between the time-series measurements at different locations, and employ kernel machines to build a predictive model. The choice of the similarity metric is critical to the performance.}}

\label{fig:overview}
\vspace{-15pt}
\end{figure*}
\subsection{Modeling Time Series Data}
Analyzing complex time series data has been a problem of significant interest in a variety of medical applications, and in neural studies in particular. In many cases, the first step is to choose an appropriate metric to summarize the overall connection between independent brain maps over time. Commonly used metrics include $\ell_2$-norm, the Pearson correlation between time series, and loadings computed based on independent component analysis. Interestingly, as we demonstrate in this paper, this metric can be the basis to construct meaningful graph representations of the brain connectivity and their local characteristics can be a powerful indicator of disease conditions. While point-wise metrics such as the $\ell_2$-norm and correlations are simple to compute, they assume the measurement at each time-step to be independent and hence cannot reveal the properties of the underlying dynamical system. In signal processing, this problem has been alleviated by constructing state-space representations for time series data. In lieu of building conventional state-space models, time delay embeddings (TDE) provide an alternative way to reconstruct the underlying dynamical system from the observed data \cite{journals/ijbc/BrownK09}. The delay embedding of a time series $\mathbf{x}$ can be defined as $\mathbf{X}_t=[x_t,x_{t+\tau},x_{t+2\tau},...,x_{t+(m-1)\tau}]$, where $m$ is the embedding dimension and $\tau$ is the delay parameter. The $m$ time-delayed observation samples can be considered as points in $\mathbb{R}^m$, which is referred to as the delay embedding space. Takens theorem ensures that for a sufficiently large $m$, delay embeddings can recover the underlying topology of the system. This has motivated the use of computational topology tools to study delay embeddings \cite{perea2015sliding}. 


%
%



\subsection{Kernel Machines}
Since kernel machines are central to the proposed system, we recap their definition. Let us consider the problem of binary classification using a Support Vector Machine (SVM) classifier that attempts to find a linear decision boundary between the two classes. When the classes are not linearly separable, it is beneficial to define a mapping function onto a high-dimensional space $\phi: \mathbb{R}^d \rightarrow \mathbb{R}^{D}$ ($D > d$), such that SVM can yield a linear decision boundary in the resulting space. It is well known that this SVM formulation can be efficiently solved by considering its Lagrangian dual based on the kernel trick \cite{thiagarajan2014multiple}. 
\begin{definition}
Given the data domain $\mathcal{X} \subset \mathbb{R}^d$, a function $k: \mathcal{X} \times
\mathcal{X} \rightarrow \mathbb{R}$ is a valid kernel if it gives rise
to a positive definite kernel matrix. i.e., $\mathbf{z}^T \mathbf{K}
\mathbf{z} \geq 0, \forall \mathbf{z} \in \mathbb{R}^d$. In addition,
a valid kernel defines an inner product and a lifting (transformation)
$\phi$, such that $k(\mathbf{x}_i,\mathbf{x}_j) = \langle
\phi(\mathbf{x}_i),\phi(\mathbf{x}_j) \rangle$ where $\langle
\cdot,\cdot\rangle$ denotes the inner product in the lifted space. This transformed space is referred as the reproducing kernel Hilbert space (RKHS).
\end{definition}In general, a kernel $k(\mathbf{x}_i, \mathbf{x}_j)$, that measures the similarity between the two samples, should be symmetric and positive semi-definite (e.g. radial basis function). Since linear operations within the RKHS can be interpreted as non-linear operations in the data domain, the linear models learned in the RKHS provide the power of a non-linear model. Another interesting property of kernel methods is that fusing multiple kernels is straightforward (e.g. convex combination), and hence multiple kernel similarities can be used jointly to improve prediction.

\section{Proposed Approach}
\label{sec:approach}
In this section, we describe the proposed approach for predicting autism severity based on activity patterns in different brain regions. Mathematically, this problem can be viewed as tensor regression \cite{zhou2013tensor}, wherein the complex relationships between the covariates, in the form of mutli-dimensional arrays, should be modeled to effectively predict the severity. Conventional approaches for tensor regression are often challenged by the ultrahigh dimensionality and complexity of structure in the data. Consequently, we adopt an alternative approach that represents each subject as a undirected graph of multiple time series measurements and builds a kernel machine to predict the dependent variable by exploiting the structural similarities between the graph representations for different subjects. Figure \ref{fig:overview} provides an illustration of the proposed system.

Graphs are natural data structures to model data in high dimensional spaces, where nodes represent the objects and the edges describe the relations between them. In our system, we construct a graph for each subject to describe the brain connectivity. For each subject, the data is collected from $K$ regions in the brain with $N$ samples each, resulting in a matrix, $F = [X_{kn}]$, where, $k \in [1,2,\dots K]$ and $n \in [1,2,\dots N]$. Since the choice of the graph construction method is critical to the success of the pipeline, we propose to explore a broad set of strategies and study their impact on the prediction performance (refer Section \ref{sec:graph}). We denote the set of graphs by $\{G_{\ell}\}_{\ell=1}^L$, where $L$ is the total number of subjects. Comparing two graphs amounts to defining a kernel that can capture the inherent structure described by the graphs, and is efficient to compute.  

A variety of graph kernels have been proposed in the machine learning literature \cite{vishwanathan2010graph} and some popular examples include the shortest path kernel and the random walk kernel. There exist several similarity measures based on graph isomorphism or related concepts such as subgraph isomorphism or the largest common subgraph. Possibly, the most natural measure of similarity is to check whether the graphs are topologically identical, i.e., isomorphic. In this work we consider the shortest path kernel and the kernel construction in \cite{wlGraphKernel} based on the Weisfeiler-Lehman test of isomorphism, which augments the node labels by the sorted set of node labels of neighboring nodes, and compresses these augmented labels into new labels. The new labels are concordant in graphs $G_i$ and $G_j$, meaning that if nodes in $G_i$ and $G_j$ have identical neighboring labels, and only in this case, they will get identical new labels. Note that with unlabeled graphs (our case), the degree of a node is used as its label. The Weisfeiler-Lehman (WL) kernel is constructed by iteratively creating multiple subgraphs using the isomorphism test and accumulating their similarities. In our experiments, we evaluate both the kernels and pick the best performing among the two. 

\section{Similarity Measures for Graph Design}
\label{sec:graph}
We use popular similarity measures, which treat the time series as feature vectors such as a PCA followed by the an RBF kernel, or aggregate temporal information such as the Pearson correlation coefficient.  In addition, we study the following strategies to construct graphs for each subject.

\noindent \textbf{$\ell_1$-graph:} Using a generative model to encode the measurement at one location using the time series data from the rest enables a sophisticated modeling of the similarities. In particular, we adopt ideas from the sparse coding literature to build an $\ell_1$ graph. In sparse coding \cite{thiagarajan2014image}, a vector is represented as a linear combination of a parsimonious set of columns from a dictionary matrix. In our case, the generative model for sparse coding of a time series at location $i$ is given by $\mathbf{x}_i = \mathbf{X}\mathbf{a}_i$, where $\mathbf{a}_i \in \mathbb{R}^K$ is the coefficient vector, which is assumed to be sparse \textit{a priori}. For each subject $i$, we solve, $\min_{\mathbf{a}_i} \|\mathbf{x}_i - \mathbf{X} \mathbf{a}_i\|_2^2 + \lambda \|\mathbf{a}_i\|_1 $ subject to $a_{ii} = 0, \mathbf{a}_i \geq \mathbf{0}$. Here, the $\|.\|_1$ denotes the $\ell_1$ norm (convex surrogate for sparsity), the $\|.\|_2$ terms measures the quality of fit. Note that the constraint $a_{ii} = 0$ ensures that $\mathbf{x}_i$ does not contribute to its own representation, and the  non-negativity constraint ensures that the graph weights are non-negative.

\noindent \textbf{Persistence Homology: }We construct the delay embedding for each time series and use persistent homology to obtain the similarity metric. Persistent homology is a method to extract topological information of point clouds or functions \cite{Edelsbrunner2002}. More specifically, \textit{Betti} numbers can be interpreted as the number of holes in each dimensions, which is the topological feature of interest. For example, given a manifold or a simplicial complex, Betti 0 be interpreted as the connected components, and Betti 1 being reflects the periodic structure in the time series. Filtration records the historical construction of simplicial complex, where for each time we can record the topological information. Given a filtration, persistence of certain features are recorded as its birth and death time \cite{Zomorodian_computingpersistent}. The length of the life from the birth to death can be interpreted as the importance of the feature. Persistence diagram is a visualization of the information obtained by plotting the birth and death times of each feature as a point in $\mathbb{R}^2$. In our implementation, we used Vietoris-Rips filtration and constructed the persistence diagrams using the DIPHA software library \cite{DIPHA}.

While standard distance metrics such as the bottleneck and p-Wasserstein exist for comparing persistence diagrams, they are computationally ineffective. This has resulted in the development of a variety of scalable metrics \cite{JMLR:v16:bubenik15a, 2015arXiv151002502C,Anirudh_2016_CVPR_Workshops, NIPS2015_5887}. We adopt the persistence scale space similarity defined by Kwitt \textit{et.al.} \cite{NIPS2015_5887} which maps the diagram to a reproducing kernel Hilbert space(RKHS) with certain boundary conditions and computes the inner product in the RKHS. 
 \begin{equation} \label{keq:3}
 \begin{aligned}
 k_{\sigma}(F,G) &= <\Phi_{\sigma}(F), \Phi_{\sigma}(G)>_{L^2(\Omega)} \\
 &= \frac{1}{8\pi\sigma} \sum_{p \in F, q \in G} e^{-\frac{||p-q||^2}{8\sigma}} - e^{-\frac{||p-\bar{q}||^2}{8\sigma}}
\end{aligned}
 \end{equation}
Then $k_{\sigma}$ is positive definite and is stable with bottleneck distance as its upper bound \cite{NIPS2015_5887}.

\section{Experiments}
\label{sec:results}
Our main goal is to predict the ADOS score, which is a measure of severity of autism. However, it is known that the ADOS score is not objective, and depends on other factors such as the subject's developmental and language level \cite{Moradi2016}. In order to work around this problem, \cite{Moradi2016} suggests an alternative score that is mapped from the ADOS score, they call a ``severity score'', that takes into account the various factors. In this work, we solve a more tractable problem, of classification instead of regression, by dividing the ADOS scores into three levels - mild (0-8), moderate (9-13) and severe (13 and above). These levels were chosen so that the three classes are equally well represented in each dataset. 

\noindent \textbf{Data}: We evaluated the proposed methods on datasets from two different sites in the ABIDE collection \cite{ABIDEpaper2014}. We treated the two datasets independently, as they are expected to come from different distributions owing to the variations in parameter settings, scanner models etc. during the acquisition process. This is similar to the protocol followed in \cite{WongISBI2016}. We used data from the ``UCLA'' and ``USM'' sites, pre-processed \footnote{\url{http://preprocessed-connectomes-project.org/abide/index.html}} using the Neuro Image Analysis Kit (NIAK) pipeline, with band pass filtering and global signal regression (`filt-global'). We selected the \textit{rois-ho} regions of interest, which had $111$ locations from which the time series were extracted. 

After selecting the subjects that have a valid ADOS score for training, the UCLA dataset contains $28$ subjects, and the USM dataset contains $58$ subjects. In all the experiments we performed a leave-one-out (LOO) training strategy, where for each sample in the dataset, we predict using a kernel SVM classifier trained on the rest.

\noindent \textbf{Graph construction:} We utilize the various graph construction strategies listed in section IV, to compute the connectivity graph for each subject. This matrix is normalized to have values between $(0,1)$ as it needs to be thresholded to construct the binary graph. We perform two sets of experiments with the similarity matrix -- First, the upper triangular part of the graph is vectorized and treated as a feature and classified using a traditional linear SVM. Next, we treat it as an adjacency matrix, which is converted to a binary graph after thresholding. The classification results for two different datasets are shown in table \ref{tab:expt}. To construct the similarity matrix based on persistence diagrams, we first fixed time delay parameters, time delay $\tau =3$, and embedding dimension $m=2$. Next we compute the Betti-1 numbers of this point cloud using DIPHA \cite{DIPHA}. Here, we are computing one persistent diagram per location, resulting in several persistent diagrams per subject, following which we used persistence scale space similarity to build the graph. The results are shown in tables \ref{tab:expt_usm},\ref{tab:expt_ucla}.

\begin{table}[t]
\centering 
\subfloat[][USM dataset with 58 subjects]{ 
\begin{tabular}{ |c|c|c| } \hline
Feature 			& Traditional Kernel& Graph Kernel 	\\ \hline
PCA 				&  29.31 & \textbf{46.42} \\ \hline
RBF		 			&  32.75 & \textbf{48.28} \\ \hline
Correlation		 	&  41.37 & \textbf{55.17} \\ \hline
$\ell_1$ graph	 	&  24.13 & \textbf{50.00} \\ \hline
Betti-1 Persistence Diagram (PD)	&  44.82 & \textbf{48.27} \\ \hline
Sum kernel (PD, $\ell_1$) &43.10& \textbf{46.55} \\\hline
\end{tabular}
\label{tab:expt_usm}
}

\subfloat[][UCLA dataset with 28 subjects]{ 
\begin{tabular}{ |c|c|c| } \hline
Feature 			& Traditional Kernel& Graph Kernel 	\\ \hline
PCA 				&  42.85 & \textbf{50.00} \\ \hline
RBF		 			&  32.14 & \textbf{39.28} \\ \hline
Correlation		 	&  39.28 & \textbf{42.85} \\ \hline
$\ell_1$ graph	 	&  28.57 & \textbf{42.85}	\\ \hline
Betti-1 Persistence Diagram (PD) & 28.57 & \textbf{42.85} \\\hline
Sum Kernel (PD, $\ell_1$) & 21.42 & \textbf{50.00} \\\hline
\end{tabular}
\label{tab:expt_ucla} 
}
 \caption{\small{Comparing classification performance on two datasets from the ABIDE data collection. We compare various graph construction strategies with traditional and graph kernels, to show there is a consistent and significant advantage towards using the latter.}}
 \label{tab:expt}
\end{table}

\section{Conclusion}
\label{sec:conclusion}
In this paper, we studied the advantages of using graph kernels for autism spectrum disorder classification. We performed experiments with the ABIDE dataset, consisting of multidimensional time series which is obtained from resting state fMRI. We showed that the graph kernel is an effective strategy across different kinds of graphs constructed using similarity measures on time series such as -- correlation, $\ell_1$ graph, persistent homology. The graph kernel with a temporal feature such as persistent homology, combines both spatial and temporal information to effectively model the dynamics of a brain network.  

\small{ 
\bibliographystyle{ieee}
\bibliography{refs}
}
\end{document}